\documentclass[runningheads]{llncs}
\usepackage{graphicx}

\usepackage{tikz}
\usepackage{comment} 
\usepackage{amsmath,amssymb} 
\usepackage{color}
\usepackage{microtype}
\usepackage{subcaption}
\newcommand{\etal}{\textit{et al.}}

\begin{document}
\pagestyle{headings}
\mainmatter
\def\ECCVSubNumber{3627}  
\title{Incorporating Reinforced Adversarial Learning in Autoregressive Image Generation} 

\titlerunning{Reinforced Adversarial Learning in Autoregressive Image Generation}
%
\author{Kenan E. Ak \inst{1} \and
Ning Xu \inst{2} \and
Zhe Lin \inst{2} \and
Yilin Wang \inst{2}}
\authorrunning{Ak K.E., Xu N., Lin Z., Wang Y.}
%
\institute{Institute for Infocomm Research, A*STAR \\
\email{kenanea@i2r.a-star.edu.sg}
\and Adobe Research \\
\email{\{nxu, zlin, yilwang\}@adobe.com}}
\maketitle

\begin{abstract}
Autoregressive models recently achieved comparable results versus state-of-the-art Generative Adversarial Networks (GANs) with the help of Vector Quantized Variational AutoEncoders (VQ-VAE). However, autoregressive models have several limitations such as exposure bias and their training objective does not guarantee visual fidelity. To address these limitations, we propose to use Reinforced Adversarial Learning (RAL) based on policy gradient optimization for autoregressive models. By applying RAL, we enable a similar process for training and testing to address the exposure bias issue. In addition, visual fidelity has been further optimized with adversarial loss inspired by their strong counterparts: GANs. Due to the slow sampling speed of autoregressive models, we propose to use partial generation for faster training. RAL also empowers the collaboration between different modules of the VQ-VAE framework. To our best knowledge, the proposed method is first to enable adversarial learning in autoregressive models for image generation. Experiments on synthetic and real-world datasets show improvements over the MLE trained models. The proposed method improves both negative log-likelihood (NLL) and Fréchet Inception Distance (FID), which indicates improvements in terms of visual quality and diversity. The proposed method achieves state-of-the-art results on Celeba for 64$\times$64 image resolution, showing promise for large scale image generation.
\end{abstract}

\begin{figure}
\centering
\includegraphics[scale=0.32]{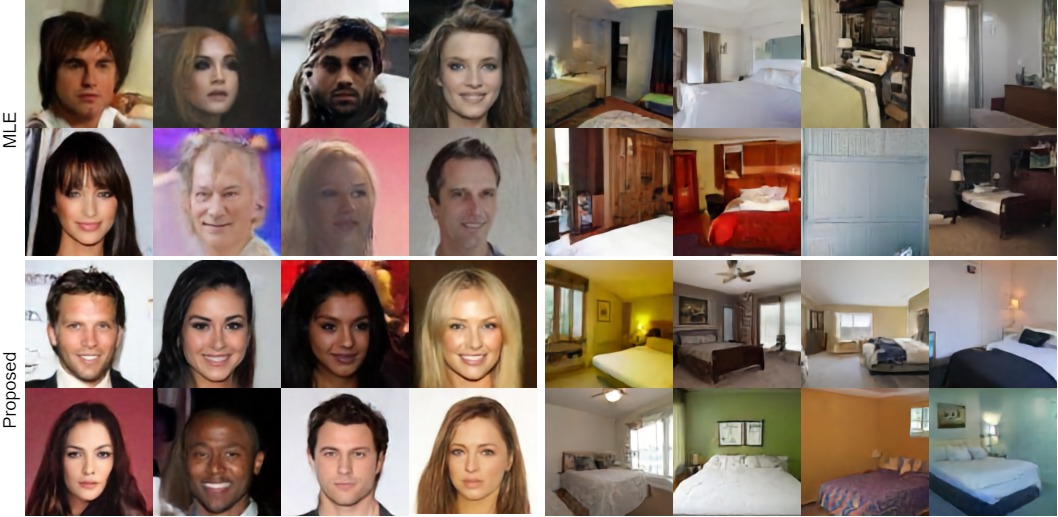}
\caption{Sample comparison of the proposed reinforced adversarial learning with the MLE trained model on CelebA~\cite{liu2015faceattributes} and LSUN-bedroom~\cite{yu15lsun} datasets. Results are randomly sampled.} \label{fig:first_fig}
\end{figure} 

\section{Introduction}
Image generation is a central problem in computer vision and has numerous applications. Nowadays, powerful image generation methods are mostly based on Generative Adversarial Networks (GANs), which was first introduced by Goodfellow~\etal~\cite{goodfellow2014generative}. With the development of advanced network structures and large-scale training~\cite{brock2018large,zhang2018self}, GANs are able to generate high-quality and high-resolution images. However, it is known that GANs do not capture the complete diversity of the true distribution~\cite{arora2017generalization,arora2018do}. Additionally, GANs are difficult to evaluate where hand-crafted metrics such as Inception Score~\cite{salimans2016improved} and Fréchet Inception Distance (FID)~\cite{heusel2017gans} must be used to test their performance.

Autoregressive models~\cite{oord2016pixel,kingma2013autoencoding,oord2016conditional} directly optimize negative log-likelihood (NLL) on training data offer another way for image generation. These models are less likely to face the mode collapse issue due to their objective~\cite{razavi2019generating}. Additionally, the objective itself provides a good evaluation metric. PixelCNN~\cite{oord2016conditional} is a common choice due to its performance and computational efficiency. The introduction of two recent works~\cite{razavi2019generating,Fauw2019HierarchicalAI}, which make use of PixelCNNs, vector quantization~\cite{van2017neural} and hierarchical-structure have shown comparable results vs. GANs. These advancements could open up a new avenue for image generation research.

On the other hand, there are several open problems in likelihood-based methods. As pointed out in~\cite{theis2015note}, optimizing NLL does not necessarily lead to generating realistic images as this objective is not a good measure of visual quality. Another issue is the exposure bias. The sampling of PixelCNN is sequential, which is a different procedure from its training stage. This discrepancy can be troublesome as small errors during the sampling can accumulate towards the next steps, which may lead to unrealistic samples~\cite{bengio2015scheduled}. To alleviate these issues, VQ-VAE-2~\cite{razavi2019generating} proposed to use the classifier based rejection sampling. This technique can eliminate low-quality samples based on an ImageNet pre-trained classifier. Classifier based rejection sampling helps VQ-VAE-2 achieve competitive results versus BigGAN~\cite{brock2018large}. Nevertheless, this method is time exhaustive, especially considering the slow sampling speed of autoregressive models. Additionally, class information might not be always available in most datasets.

In this paper, we aim to leverage both advantages of GANs and likelihood models. Consequently, our objective is to further improve the visual fidelity of autoregressive models while addressing the exposure bias issue. We focus on the VQ-VAE framework~\cite{van2017neural,razavi2019generating}, which relies upon PixelCNN~\cite{oord2016conditional} for latent code generation and uses a pre-trained decoder to reconstruct images. However, PixelCNN~\cite{oord2016conditional} generates sequences of discrete codes. Therefore, we cannot directly optimize it directly with GAN objectives. To fix this issue, we regard the sampling process of PixelCNN as a sequential decision-making process and optimize it by Reinforced Adversarial Learning (RAL). A discriminator network is trained to distinguish between real \& fake images and provides rewards to the PixelCNN. We compare our proposed method on two image generation benchmarks to demonstrate that RAL can indeed improve both FID and NLL significantly.

Adversarial training for sequence generation with RL is first explored in SeqGAN~\cite{yu2016seqgan} but there is a lot of room for improvement, especially for images. Compared to text or music generation tasks studied in SeqGAN, image generation is more complex due to longer sequences and strong spatial correlation of images. In this paper, in addition to applying RAL work for image generation for the first time, we also propose a partial generation idea to address the issue with long-sequences. Moreover, we include a method that enables the use of intermediate rewards for spatial correlation. The proposed method also enables the incorporation of independently trained modules in VQ-VAE. Our experiments show that RAL can greatly improve one of the most successful image generation methods for autoregressive models, which has a lot of potential for future work. \\

In summary, our contributions are four-fold:

\begin{itemize}
    \item We propose to augment the autoregressive image generation with adversarial training by using reinforcement learning to further improve the image quality while incorporating collaboration between independently trained modules.
    \item We utilize a patch-based discriminator to enable intermediate rewards.
    \item We propose partial generation for faster training to addresses the sampling issue of long sequences.
    \item We conduct extensive experiments to show that the proposed method significantly improves over the MLE trained baselines in different settings for both synthetic and real datasets.
\end{itemize}

The rest of the paper is organized as follows. In Sect.~\ref{sect: Related}, we briefly introduce related works. In Sect.~\ref{sect: VQ-VAE-2}, we introduce the basics of VQ-VAE \& VQ-VAE-2 and we describe our method in detail in Sect.~\ref{sect: Method}. Implementation details are provided in Sect.~\ref{sect: Implementation}, experimental results are presented in Sect.~\ref{sect: Experiments} and finally, we conclude the paper in Sect.~\ref{sect: Conclusion}.

\section{Related Work} \label{sect: Related}
\subsection{Generative Models} 
GANs~\cite{goodfellow2014generative} and conditional-GANs~\cite{mirza2014conditional} have shown a huge success for many problems in computer vision~\cite{choi2017stargan,ak2019gan,ak2020semantically,emir2019semantically,Tao18attngan,kenan_ICIP,heqing_ICIP,ak2019deep} and speech processing~\cite{saito2017statistical,sisman2019study,sisman2019singan,sisman2018adaptive,sisman2020generative,zhou2020transforming}. Advanced architectures such as SaGAN~\cite{zhang2018self}, StyleGAN~\cite{karras2019style} and BigGAN~\cite{brock2018large} have shown GANs' superiority in terms of image quality compared to Variational Autoencoder (VAE)~\cite{kingma2013autoencoding} and autoregressive models, e.g., PixelCNN~\cite{oord2016conditional}. Additionally, GANs are much faster in inference compared to autoregressive models. However, it is known that GANs models can not fully capture the data distribution and may sometimes suddenly drop modes~\cite{arora2017generalization,arora2018do}. Moreover, there are no perfect metrics to evaluate the sampling quality of GANs.

Autoregressive models are less likely to face the mode collapse and can be easily evaluated by measuring log-likelihood~\cite{razavi2019generating}. However, these models are expensive to train and extremely slow at the sampling time when trained in the pixel space. Performing vector quantization~\cite{van2017neural,chorowski2019unsupervised,tjandra2019vqvae} and training a PixelCNN prior in discrete latent space is much more efficient than pixel space, which enables the scaling to large resolution images. Following this idea, the recently introduced VQ-VAE-2 framework managed to achieve competitive results compared to state-of-the-art GANs. Concurrently with VQ-VAE-2, Fauw et al.~\cite{Fauw2019HierarchicalAI} followed a similar approach to combine autoencoders with autoregressive decoders. The advantage of VQ-VAE-2 is the feed-forward design of the decoder, making it faster for sampling. Due to faster sampling time, VQ-VAE-2 is more feasible for our work, which requires sampling of numerous images during the training.

Although these aforementioned autoregressive models achieve decent improvements in terms of image coherence and fidelity than before~\cite{oord2016conditional,chen2017pixelsnail}, GANs are still preferred due to sampling speed and image quality. Additionally, as mentioned by Theis et. al~\cite{theis2015note}, good performance on log-likelihood training objective does not always guarantee good samples. In order to reduce this gap, VQ-VAE-2 proposes to use classifier-based rejection sampling to trade-off diversity and quality inspired by BigGAN~\cite{brock2018large}. However, this approach requires the autoregressive prior to sample many latent codes, which is time exhaustive and not ideal for real-world problems. Additionally, the rejection sampling method relies on a ImageNet pretrained classifier and might not be available for another dataset such as CelebA~\cite{liu2015faceattributes} and LSUN-bedroom~\cite{yu15lsun}. Lastly, the rejection sampling method decreases diversity in image samples. We propose to use reinforcement learning to improve the sampling quality of autoregressive models and enable a similar mechanism for training and testing.

\subsection{Reinforcement Learning in Sequence Generation}
RL for sequence generation has been applied by~\cite{Ranzato2015SequenceLT}, which uses bilingual evaluation understudy (BLEU)~\cite{papineni2002bleu} as a reward function to guide the generation process. SeqGAN~\cite{yu2016seqgan} is the first to train a discriminator through adversarial learning and use reinforcement learning based on policy gradient~\cite{sutton2000policy} to provide rewards for seq-to-seq network~\cite{SutskeverVL14}. Recent works~\cite{Ganin2018spiral,huang2019learning} also showed that adversarial training can be used to iteratively update images where rewards are estimated by a discriminator network. In this work, we similarly employ a discriminator network to provide rewards for the generated samples and improve sampling quality. We enable intermediate rewards by using a PatchGAN~\cite{pix2pix2017} network and use a partial generation procedure to handle long sequences.
 
\section{Background: VQ-VAE \& VQ-VAE-2} \label{sect: VQ-VAE-2}
Our method is built upon VQ-VAE~\cite{van2017neural} \& VQ-VAE-2~\cite{razavi2019generating} frameworks that are trained in two stages as follows:

\textbf{Stage 1: Learning Hierarchical Latent Codes.} In this stage, VQ-VAE learns an \textit{encoder} ${E}$ that transforms an image into a set of features, a \textit{codebook} ${C}$ that maps the real-valued features into a set of discrete latent codes and a \textit{decoder} ${U}$ that reconstructs the image from these latent codes. $E$, $C$ and $U$ are learned together as in~\cite{van2017neural}. However, the reconstruction in VQ-VAE is not perfect due to the quantization, therefore VQ-VAE-2~\cite{razavi2019generating} proposed to learn hierarchical latent codes to alleviate this problem. In VQ-VAE-2, the first hierarchy called top-level latent codes captures global image information such as shape and structures while the other hierarchy called bottom-level latent captures fine-grained details. Once stage 1 is learned, an input image can be represented as the hierarchical latent codes for compression and reconstructed by the decoder.

\textbf{Stage 2: Learning Priors Over Latent Codes.}
PixelCNN~\cite{oord2016conditional} is an autoregressive model and has been previously applied to learn image distributions in the pixel space. In VQ-VAE, PixelCNN is used to learn priors in the latent space, which has much smaller dimensionality than the pixel space. For the two hierarchies in VQ-VAE-2, PixelSNAIL~\cite{chen2017pixelsnail} is used to learn the top-level latent codes while a lighter PixelCNN is used to learn the bottom-level as the size of the bottom-level latent codes is larger. These two models are trained separately by MLE. After the training, latent codes sampled from the PixelCNN priors can be reconstructed with the pretrained decoder to an image.

\section{Reinforced Adversarial Learning} \label{sect: Method}
There are several drawbacks of PixelCNNs priors used in the VQ-VAE frameworks~\cite{van2017neural,razavi2019generating}. First, they are trained with MLE which is not a good measure of sampling quality~\cite{theis2015note}. Second, during the training, all latent codes are estimated from the real images while during inference each code is sampled, which could never be observed during the MLE training. Such discrepancy could result in an unrealistic generation. Lastly, each hierarchy and the decoder in the VQ-VAE-2 framework are trained separately, which may not collaborate well when put together for sampling images.

\begin{figure*}[t]
\centering
\includegraphics[scale=0.8]{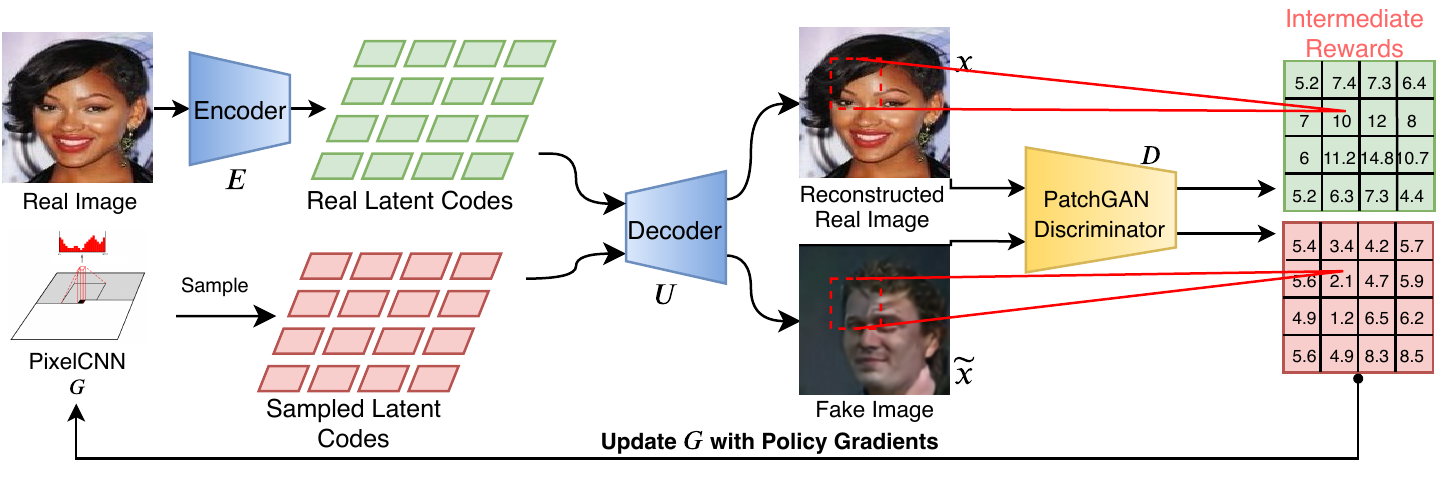}
\caption{Overview of the proposed Reinforced Adversarial Learning (RAL) framework. Initially, PixelCNN $G$ is used to sample fake latent codes which are then fed into Decoder to reconstruct the fake image. Similarly, the real image can be reconstructed from the real latent codes. We use PatchGAN Discriminator to provide rewards for overlapping regions in the image. These rewards can then be used to update $G$ with policy gradients.} 
\label{fig:network}
\end{figure*} 

In this section, we describe how we introduce adversarial learning from GANs into likelihood models to leverage the best from both worlds. Our motivation is to let the PixelCNNs be able to generate samples that can fool the discriminator that is trained to distinguish between real and fake images. In this way, our PixelCNNs are trained to generate realistic sequences and the training process is exactly the same as inference. In addition, the harmony of different PixelCNNs and the decoder are also improved since they are optimized together. One issue is that PixelCNN cannot be directly optimized by the adversarial loss, which we solve by employing reinforcement learning. Parameters of Encoder \& Decoder are fixed since including them to the training corrupts the image decoding procedure, which results in having poor quality of images even when using real latent maps. 

We illustrate our solution in Sect.~\ref{sect:policy}. Our patch-based discriminator and reward definition are presented in Sect.~\ref{sect:disc}. We describe the idea of partial generation that is useful for the generation of large images in Sect.~\ref{sect:partial}. Training details are given in Sect.~\ref{sect:training}. An overview of the proposed RAL framework is shown in Figure \ref{fig:network}.

\subsection{Policy Gradients}\label{sect:policy}
For simplicity, let us first consider learning a single PixelCNN $G$ on one hierarchy of latent codes. The extension to multiple PixelCNNs on multiple hierarchies (VQ-VAE-2) is straightforward.

In adversarial training, a generator is directly optimized to maximally confuse a discriminator. However, at each time $t\in\left[0,T-1\right]$, our PixelCNN generates a discrete code $c_t$ based on the conditional probability $G(c_t|c_0,...,c_{t-1})$ given all the previous codes $(c_0,...,c_{t-1})$. This process is non-differentiable and thus cannot be directly optimized. However it can be regarded as a decision making process where the state $s_t=(c_0,...,c_{t-1})$ and the action $a_t=c_t$. We use policy gradients~\cite{sutton2000policy} to solve this problem where the objective is: 
\begin{equation} \label{eq:1}
J_G = \mathbf{E}_{a \sim G} [R_T]
\end{equation} 
where $R_T$ is the reward for the whole sequence of latent codes generated by $G$. The gradients of Equation~\ref{eq:1} can be defined as:
\begin{equation}
\nabla_{\theta} J_G\propto\mathbf{E}_{a_t \sim G}\left[ \nabla_{\theta} \log G(a_t|s_t)Q(s_t,a_t)\right]
\end{equation}
where ${\theta}$ is the parameters of the PixelCNN. $Q(s_t,a_t)$ is the action-state value and is defined as:
\begin{equation}
Q(s_t,a_t) = \sum_{k=t+1}^T\gamma^{k-t-1}r_k
\end{equation}
where $r_k$ is the reward at time step $k$, which can be obtained by the discriminator and will be discussed later. $\gamma$ is the discounted factor within the range of $\left[0,1\right]$. 

We use the REINFORCE~\cite{williams1992simple} algorithm to roll-out the whole sequence after $a_t$ is sampled by using the same PixelCNN $G$. In our experiment, we only do one Monte-Carlo roll-out for the training-speed concern. Finally, $G$ can be updated as follows where the $\alpha$ is learning rate:
\begin{equation}
\theta \leftarrow \theta+\alpha \nabla_{\theta} J_G
\end{equation}
\subsection{Discriminator} \label{sect:disc}
After a sequence of latent codes $(c_0,...,c_{T-1})$ is generated, the decoder $U$ is used to reconstruct the image $\mathbf{\tilde x}$, as shown in Figure~\ref{fig:network}. Then a discriminator $D$ is trained to distinguish between the generated and real images (real images are also their reconstructed version). We use the WGAN loss~\cite{arjovsky2017wasserstein,gulrajani2017improved} instead of the original GAN loss proposed in~\cite{goodfellow2014generative}. Since it provides smoother gradients which enable the generator to still learn even when the discriminator is performing strong. The loss function of our discriminator is defined as:
\begin{equation}
\begin{array}{l}
L_{D}  =- \mathbf{E}_{\mathbf{x} \sim p_{d}}[D(\mathbf{x})]  + \mathbf{E}_{\mathbf{\tilde x} \sim G}[D(\mathbf{\tilde x})] +  \\ \lambda_{gp} \mathbf{E}_{\mathbf{\hat{x}} }[(|| \bigtriangledown_{\mathbf{\hat{x}}} D(\mathbf{\hat{x}}))||_2-1)^2]
\end{array}\label{eq:6}
\end{equation}
where $p_d$ is the distribution of real images. The final term is the gradient penalty weighted by $\lambda_{gp}$. $\mathbf{\hat{x}}$ is an image sampled uniformly along a straight line between real and generated images. 

Our discriminator follows a similar structure to the PatchGAN discriminator~\cite{pix2pix2017}. In contrast to traditional GAN discriminators that only produce a single scalar output, the PatchGAN discriminator can provide a score map $S$, each element of which corresponds to the score of a local image patch as shown in Figure~\ref{fig:network}. There are several advantages of using this type of discriminator. First, it provides a good measure of the realism of local patches. Second, the scores can be used as intermediate rewards which can alleviate the issue of sparse rewards in RL training. Third, its fully convolutional structure can handle arbitrary image sizes, which is convenient for our partial generation.

The \textit{intermediate reward} at every time step can be computed by upsampling the score map $S$ to the original size of the latent codemap as shown in Figure~\ref{fig:network}. The reward $r_{t+1}$ of the sampled code $c_t$ has a corresponding location at the upsampled score map and can be easily obtained. Since each code is contained in a local region, the action-state value function $Q(s_t,a_t)$ thus more focus on the local realism.

Our \textit{single reward} function is defined as:
\begin{equation}
r_{t}=\left\{\begin{array}{ll}{0,} & {t<T} \\ {D\left(\mathbf{\tilde{x}}\right),} & {t=T}\end{array}\right.
\end{equation}
where $D\left(\mathbf{\tilde{x}}\right)$ is the average value of the score map $S$. In this case the action-state value function $Q(s_t,a_t)$ focus more on the long-term reward, i.e. the realism of the whole image.

\subsection{Partial Generation}\label{sect:partial}
\begin{figure}[t]
\centering
\includegraphics[scale=0.36]{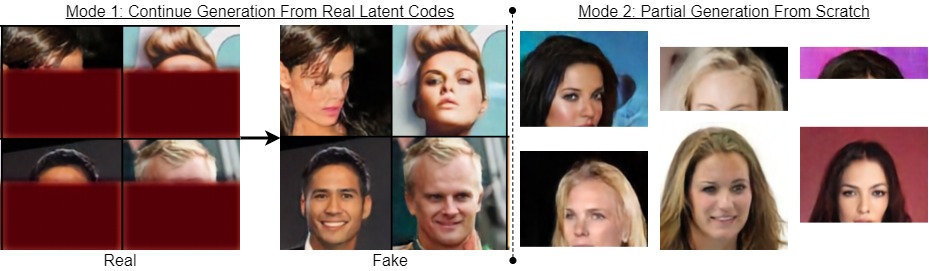}
\caption{Examples for partial generation from $G$. The proposed idea, improves the training speed of RL algorithm.} \label{fig:partial}
\end{figure}

In order to generate large-size images, the length of sampled latent codes also needs to be increased, which is time-exhaustive. This is especially troublesome as our RL training requires many Monte-Carlo roll-outs at each iteration. As a solution, we propose to use partial generation whose central idea is to only sample partial images, which can already provide meaningful rewards while improving sampling efficiency greatly. 

Specifically, the partial generation works under two modes: (1) continue generation from real latent codes and (2) partial generation from scratch. During each roll-out in RL training, the algorithm first randomly decides one mode. If mode-1 is selected, then a random number representing the number of rows of real latent codes is chosen. Then following these real codes, $G$ samples the rest of the sequence. If mode-2 is selected, a random number representing the number of rows of fake latent codes is chosen. Then $G$ samples this number of codes from scratch. The two modes are essential and improve the generation of the PixelCNN on different image regions. In addition, mode-1 provides real latent codes as context and is useful for the image completion task. Figure~\ref{fig:partial} shows some sampled examples of both the two modes. It should be noted that although the sampled codes correspond to different image sizes, as our discriminator is fully convolutional and thus has no issue.

\subsection{Training}\label{sect:training}
VQ-VAE is first trained with the two stages as described in Sect.~\ref{sect: VQ-VAE-2} and PixelCNN is pretrained with the MLE objective. Then our RL training starts. We use Adam optimizer~\cite{kingma2014adam} with $\beta_1$ = 0.5, $\beta_2$ = 0.999 and set the learning rate of the discriminator and PixelCNN to 1e-4, 4e-6 respectively with the mini-batch size of 16. Initially, the discriminator is trained for 100 iterations to catch up with the pretrained PixelCNN. During the RL training, for each PixelCNN update, the discriminator is updated 5 times and updates are performed iteratively. 
When there are two hierarchies of latent codes available, both the PixelCNN models are trained at the same time which encourages collaboration between them. Outputs from the discriminator are first normalized to the range of (-1, 1) before updating the PixelCNNs. The normalization is performed with respect to the highest value from a set of fake and real images. Without normalizing the discriminator's outputs, rewards become unreliable as, after each update, the discriminator's outputs may change drastically even with the WGAN penalty. By normalizing rewards based on the maximum value of real and fake images, rewards act as an evaluation score from the current state of $D$. For the partial generation, the algorithm randomly switches between different modes and samples a random number of rows. We set $\lambda_{gp}$ in Eq.~\ref{eq:6} to $10$ and the discounted factor $\gamma$ to $0.99$.


\section{Experiments} \label{sect: Experiments}
We first conduct experiments on a synthetic dataset similarly to~\cite{yu2016seqgan,toyama2018toward} in Sect.~\ref{sect:oracle-exp}. The synthetic dataset is constructed from a pre-trained PixelCNN~\cite{oord2016conditional} prior, which we denote as the oracle model $G_{\text{oracle}}$. The oracle model is used to provide the true data distribution and the generator is trained to fit the oracle distribution. The advantage of having an oracle model is that the generated samples can be evaluated with NLL which is not possible for real data. 

Our second experiment in Sect. \ref{sect:real-exp} is for real-world images where we train our models in the CelebA dataset~\cite{liu2015faceattributes} with different scales:  $64 \times 64$, $128 \times 128$ and $256 \times 256$, in addition to the LSUN-bedroom~\cite{yu15lsun}. As an evaluation metric, we use Fréchet Inception Distance (FID)~\cite{heusel2017gans}, which uses an Inception network to extract features to compare the closeness of real and fake images statistics. Lower FID means better image quality and diversity. Several experiments with different settings and architecture details are included in the supplementary material.

We also perform an ablation study in Sect. \ref{sect:ablation} followed up by a use-case of the proposed method for image completion task in Sect. \ref{sect:completion}, which also demonstrates the proposed method captures global structure better than MLE trained model as the generated sequences are more correlated with the real ones.


\subsection{Implementation Details} \label{sect: Implementation}
This section describes the implementation details of different modules used in RAL. The detailed information on architecture choices is included in our supplementary material.

\textbf{Encoder. }VQ-VAE~\cite{van2017neural} encoder is used to compress images to latent space. Each layer includes a transposed convolution for downsampling. For hierarchical codes, two latent levels are used i.e., top and bottom. Images are encoded into following latents codes for different resolutions: a) 64$\times$64 $\rightarrow$ 8$\times$8, b) 128$\times$128 $\rightarrow$ 16$\times$16, 32$\times$32, c) 256$\times$256 $\rightarrow$ 32$\times$32, 64$\times$64.

\textbf{Decoder. }VQ-VAE decoder is used to reconstruct images from latent codes. For hierarchical codes, the bottom-level code is conditioned on the top level. The network consists of transposed convolutional layers for upsampling.

\textbf{PixelCNN/PixelSNAIL. }Similar to VQ-VAE-2, we use PixelCNN with self-attention layers to model top-priors. For bottom-priors, which is more computationally expensive, we remove self-attention layers and reduce the number of residual channels. To improve the sampling speed from prior networks, we use caching similar to~\cite{ramachandran2017fast} and avoid redundant computation. Note that, we use a smaller VQ-VAE-2 architecture compared to~\cite{razavi2019generating} due to computational limitations. The main issue is the model capacity requirement for MLE pre-training. Even with 8$\times$V100 GPUs (8x16GB), which is the best machine we can get, we were barely able to use similar networks designed in the VQ-VAE-2 for 128$\times$128, but not enough for 256$\times$256 due to the increase in latent code sizes.

\textbf{PatchGAN Discriminator.} The discriminator has a similar convolutional architecture to the encoder network. while at the patch level to classify whether a patch is real or fake. The discriminator has $5$ strided convolutions which output a $4 \times 4$ reward map from a $128 \times 128$ image. We also experiment with different output scales in our ablation study to find the optimum architecture.

\subsection{Synthetic Experiments} \label{sect:oracle-exp}
We first perform synthetic experiments on the CelebA dataset with the image resolution of $64 \times 64$, which are mapped into one level of $8 \times 8$ latent codes by the encoder. Instead of directly training on real images, we first train a PixelCNN $G_{\text{oracle}}$ on the $8 \times 8$ latent codes. Then this model is used as an oracle data generator to provide training data. The advantage is that we can have accurate NLL estimation, which cannot be achieved with real images since the oracle generator for real images is unknown.

Next, a lighter PixelCNN\footnote{We use a lighter PixelCNN so that the MLE model cannot be easily trained towards the oracle model.} $G_{\text{MLE}}$ is trained on training samples generated by $G_{\text{oracle}}$ with the MLE objective, which is the VQ-VAE baseline. Our method fine tunes $G_{\text{MLE}}$ with our RL training to estimate another model $G_{\text{RL}}$. For this experiment we do not use partial generation since the size of latent codes is already small. The ground truth NLL estimation of a generator $G$ can be computed as:
\begin{figure}[t]
\centering
\begin{subfigure}{.5\textwidth}
  \centering
  \includegraphics[width=0.8\linewidth]{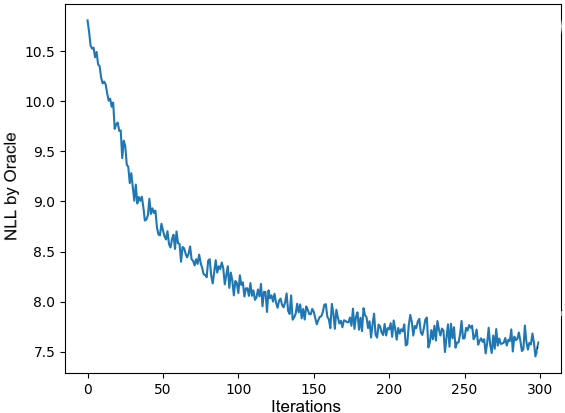}
  \caption{MLE Training}
\end{subfigure}%
\begin{subfigure}{.5\textwidth}
  \centering
  \includegraphics[width=0.8\linewidth]{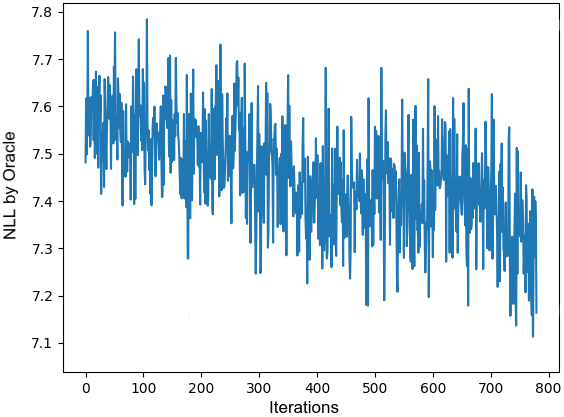}
  \caption{Proposed RL Training}
\end{subfigure}
\caption{Negative log-likelihood during the training iterations.}
\label{fig:mle_rl}
\end{figure}

\begin{equation}\label{Eq:NLL}
\mathrm{NLL}_{\text {oracle}}=-\mathbb{E}_{c_t \sim G}\left[\sum_{t=0}^{T-1} \log G_{\text{oracle}}\left(c_{t} | c_{0:t-1}\right)\right]
\end{equation}

\begin{table}[t]
\setlength{\tabcolsep}{7pt}
\caption{NLL and FID values for Oracle Experiments.}
\centering
\begin{tabular}{ c c c c }
\hline
 & NLL & FID on Oracle Data & FID on Real Data \\ \hline
MLE & 7.55 & 7.49  & 15.63 \\ 
RL - Single Reward & 7.07 & \textbf{5.17} & 14.28 \\ 
RL - Intermediate Reward & \textbf{7.05} & 5.62 & \textbf{13.69}  \\ \hline
\end{tabular}
\label{table:FID_oracle}
\end{table}

\begin{table*}[t]
\setlength{\tabcolsep}{4pt}
\caption{FID values for real-world experiments. The number in parenthesis is FID on reconstructed real images.}
\centering
\begin{tabular}{ c c c c c}
\hline
Dataset  &  \begin{tabular}[c]{@{}c@{}}CelebA\\ $64 \times 64$\end{tabular} & \begin{tabular}[c]{@{}c@{}}CelebA\\ $128 \times 128$\end{tabular}  &  \begin{tabular}[c]{@{}c@{}}CelebA\\ $256 \times 256$\end{tabular} & \begin{tabular}[c]{@{}c@{}}LSUN-bedroom\\ $128 \times 128$\end{tabular} \\ \hline
MLE & 5.35  &  58.07 (21.28) & 69.37 (55.11) & 47.77 (23.14) \\ 
Single Reward & \textbf{3.24} & \textbf{49.24 (16.49)}  & 66.44 (53.54) & 39.86 (19.37) \\ 
Intermediate Reward & 4.10  & 51.91 (19.68)  & \textbf{64.72 (51.38)} & \textbf{36.47 (16.06)} \\ \hline 
PGGAN~\cite{karras2017progressive}& - & 7.30 & - & 8.34 | $256 \times 256$\\ 
COCO-GAN~\cite{lin2019coco}  & 4.00  & 5.74  & - & 5.99 | $256 \times 256$\\ \hline
\end{tabular}
\label{table:FID_real}
\end{table*}

We report the NLL results in Table~\ref{table:FID_oracle} and the learning curves in Figure~\ref{fig:mle_rl}. It can be seen that our RAL with different rewards improve the MLE-trained model even though our objective is not NLL. 
In Table~\ref{table:FID_oracle}, we also show that our methods get better FID scores when compared with both fake images produced by $G_{\text{oracle}}$ and real images.

\begin{figure*}[t]
\centering
\includegraphics[scale=0.32]{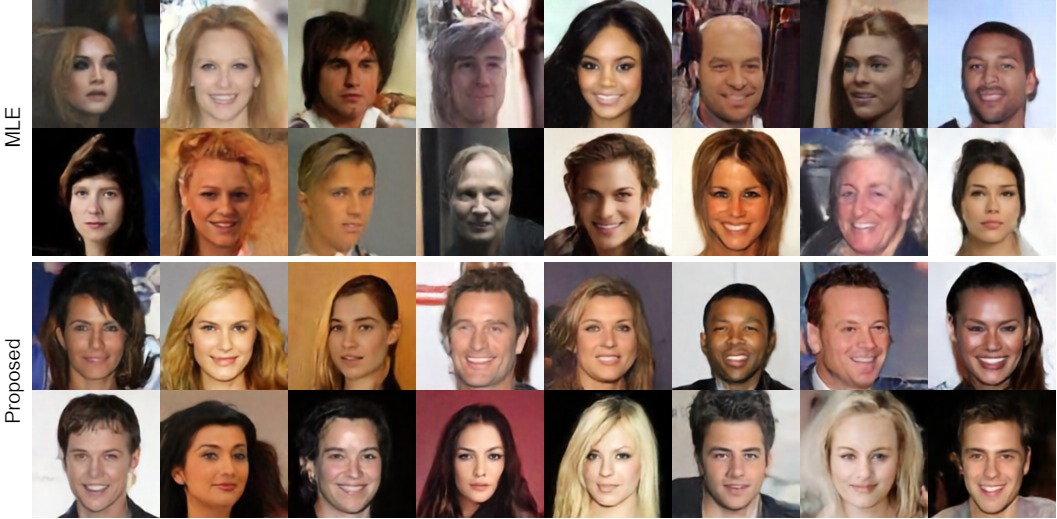}
\caption{Image samples on CelebA dataset with  $128  \times 128$ resolution. The ratio of good samples from MLE model is less than ours.}
\label{fig:celeba_128}
\end{figure*} 

\subsection{Real World Experiments}\label{sect:real-exp}
Given the success of oracle experiments, we further evaluate our method for real-word image generation and report the FID values~\cite{heusel2017gans} in Table~\ref{table:FID_real}. Note that the FID scores outside/inside parenthesis are computed between generated images and the real images/reconstructed real images. The large FID scores on $128 \times 128$ images are mainly caused by the lossy compression of the VQ-VAE encoder-decoder, which is also mentioned by~\cite{razavi2019generating}. In addition, only reconstructed real images are used during our training. Therefore the scores inside the parenthesis are more meaningful for comparison.

\begin{figure*}[t]
\centering
\includegraphics[scale=0.32]{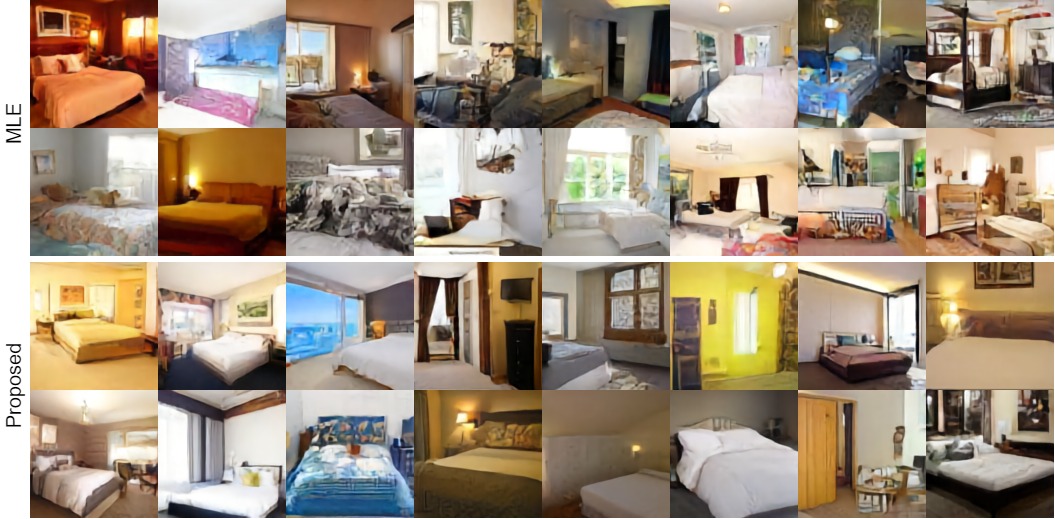}
\caption{Image samples on LSUN-bedroom dataset with  $128  \times 128$ resolution. The ratio of good samples from MLE model is less than ours.}
\label{fig:lsun_128}
\end{figure*}

Results show that our models improve the MLE trained model in all settings by a large margin. In addition, our single-reward model achieves the best FID score on CelebA $64 \times 64$ resolution even when compared to the state-of-the-art GAN models~\cite{karras2017progressive,lin2019coco} (the compression loss is relatively small on $64 \times 64$ resolution and thus our FID score is not affected much). For $128 \times 128$ resolution, the use of intermediate reward performs better on the LSUN-bedroom while the single reward is better for CelebA. We use a much smaller architecture than the one reported in~\cite{razavi2019generating} for CelebA $256  \times 256$ experiments. The proposed RAL still improves the MLE trained model which demonstrates the effectiveness of the proposed method for larger latent codes.

We provide qualitative visual comparisons with the MLE trained model for the Celeba dataset in Figure \ref{fig:celeba_128} as well as LSUN-bedroom dataset in Figure \ref{fig:lsun_128}. As can be seen, the main problem of the MLE trained model is that it generates visually good and bad images from the same model. This is mostly due to accumulating errors during the sampling and mismatch of training/testing procedures. Also, note that we do not use any classifier-based rejection sampling proposed in~\cite{razavi2019generating}. However, as the MLE trained models is additionally optimized via GAN loss, better-quality samples are produced.

\begin{table}[t]
\setlength{\tabcolsep}{1pt}
\vspace{-3mm}
    \caption{Ablation experiments on CelebA with $128 \times 128$ resolution. The number in parenthesis is FID score on reconstructed real images.}
\vspace{-3mm}
    \begin{subtable}{.5\linewidth}
      \centering
        \caption{}
        \scalebox{0.9}{
        \begin{tabular}{ l c }
 & FID \\ \hline
Partial-Gen. (Proposed) & \textbf{49.24 (16.54)}  \\ 
Full-Gen. & 51.91 (20.25) \\ \hline
RAL on Top &  54.82 (19.20) \\ 
RAL on Bottom &  55.02 (19.80) \\ \hline 
        \end{tabular}}
    \end{subtable}%
    \vspace{-3mm}
    \begin{subtable}{.5\linewidth}
      \centering
        \caption{}
        \scalebox{0.88}{
        \begin{tabular}{l c}
 & FID \\ \hline
Single reward, D output: 8 $\times$ 8 &  53.15 (18.28)  \\ 
Single reward, D output: 4 $\times$ 4 &  \textbf{49.24 (16.54)}  \\
Single reward, D output: 2 $\times$ 2 &  49.78 (17.68)   \\ \hline
Interm. reward, D output: 8 $\times$ 8 &  50.22 (18.65)  \\ 
Interm. reward, D output: 4 $\times$ 4 &  54.835 (19.47)  \\ 
Interm. reward, D output: 2 $\times$ 2 &  55.37 (20.99)   \\ \hline
        \end{tabular}}
    \end{subtable} 
    \label{table:Ablation}
\end{table}

\begin{figure*}
\centering
\includegraphics[scale=0.32]{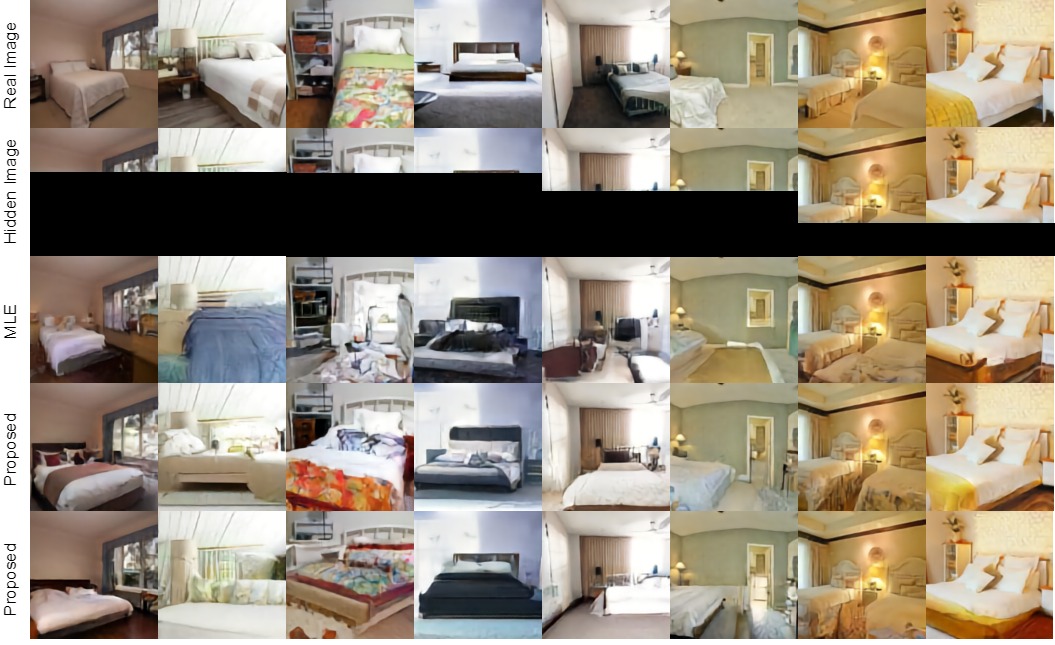}
\caption{Image completion experiment for the MLE trained method versus the proposed method. Images in the first row are hidden with masking function in the second. The generation shown in rows 3, 4 and 5 continue from visible regions of the image and the corresponding location can be approximated by the encoder. 
}
\label{fig:completion}
\end{figure*} 

\subsection{Ablation Study} \label{sect:ablation}
In Table \ref{table:Ablation}, we investigate different modules of the proposed method using the CelebA dataset with the image resolution of  $128 \times 128$. In Table \ref{table:Ablation}(a), the use of partial generation outperforms the full-generation. Note that both configurations are trained for the same time interval for a fair comparison. For all settings in Table \ref{table:Ablation} (a), we used single-reward, estimated from a discriminator, which outputs a $4 \times 4$ output. In Table \ref{table:Ablation}(a), we also test the effect of applying RAL on top and bottom priors independently, which achieves worse FID than the proposed joint training. This presents that the proposed RAL indeed enhances the collaboration of independently MLE trained priors.

Secondly, we test single and intermediate reward configurations with different score outputs from the discriminator in Table \ref{table:Ablation}(b). The score map is adjusted by changing the number of convolutional layers. The single reward works best with the score map of $4 \times 4$. Intermediate reward works best with an $8 \times 8$ score map. We also tried using larger score map outputs but did not observe improvements.

\subsection{Image Completion} \label{sect:completion}
Image completion/inpainting~\cite{bertalmio2000image,criminisi2003object} is a process of restoring missing/damaged parts of the image. As the generation of autoregressive models is sequential, i.e., the generation of the current pixel is conditioned on the previous pixels, autoregressive models can be used for image completion. As in most image generation problems, GANs are top-performing models for this task~\cite{fawzi2016image,yu2019free} but we believe this is a good application for comparisons with the MLE model. 

In Figure \ref{fig:completion}, we perform some qualitative results with the LSUN-bedroom ($128 \times 128$) dataset with different image completion settings. Firstly, a number of rows of input images (first row) are hidden (second). These hidden rows can be simply eliminated by zeroing out their corresponding locations in latent maps. Next, autoregressive models can continue sampling the remaining codes where completed reconstructed images are shown in the last three rows. We illustrate two modes of the proposed model to show the proposed method can generate diverse samples.

In the first column, a glimpse of a room is given where both methods successfully include the window in the generated image. The proposed method tends to produce more realistic samples due to its adversarial learning. Another interesting example can be seen in the final column, where the proposed method can reproduce the remaining regions more realistically compared to the MLE trained model. 

\section{Conclusion} \label{sect: Conclusion}
Our proposed idea has several advantages compared to traditional autoregressive models. By utilizing reinforcement learning, we bridge the gap between training and testing procedures while enabling the autoregressive model to take advantage of GAN training. Additionally, we use partial generation to improve the training and use two different image rewards. Last but not least, we show that the proposed method can improve the collaboration of independently trained hierarchical modules. To our best knowledge, our framework is first to enable the adversarial learning in PixelCNNs and does not solely depend on the traditional objective function of autoregressive models for good sample quality. A possible future direction would be to extend the proposed framework to higher resolution images which would be possible with faster sampling times.

\clearpage
%
%
\bibliographystyle{splncs04}
\bibliography{MyCollection}

\begin{thebibliography}{10}
\providecommand{\url}[1]{\texttt{#1}}
\providecommand{\urlprefix}{URL }
\providecommand{\doi}[1]{https://doi.org/#1}

\bibitem{ak2019deep}
Ak, K.E.: Deep Learning Approaches for Attribute Manipulation and Text-to-Image
  Synthesis. Ph.D. thesis (2019)

\bibitem{ak2019gan}
Ak, K.E., Lim, J.H., Tham, J.Y., Kassim, A.A.: Attribute manipulation
  generative adversarial networks for fashion images. In: ICCV. IEEE (2019)

\bibitem{emir2019semantically}
Ak, K.E., Lim, J.H., Tham, J.Y., Kassim, A.A.: Semantically consistent
  hierarchical text to fashion image synthesis with an enhanced-attentional
  generative adversarial network. In: ICCVW (2019)

\bibitem{ak2020semantically}
Ak, K.E., Lim, J.H., Tham, J.Y., Kassim, A.A.: Semantically consistent text to
  fashion image synthesis with an enhanced attentional generative adversarial
  network. PRL  (2020)

\bibitem{kenan_ICIP}
Ak, K.E., Ying, S., Lim, J.H.: Learning cross-modal representations for
  language-based image manipulation. In: ICIP (2020)

\bibitem{arjovsky2017wasserstein}
Arjovsky, M., Chintala, S., Bottou, L.: Wasserstein generative adversarial
  networks. In: ICML. pp. 214--223 (2017)

\bibitem{arora2017generalization}
Arora, S., Ge, R., Liang, Y., Ma, T., Zhang, Y.: Generalization and equilibrium
  in generative adversarial nets (gans). In: ICML. pp. 224--232 (2017)

\bibitem{arora2018do}
Arora, S., Risteski, A., Zhang, Y.: Do {GAN}s learn the distribution? some
  theory and empirics. In: International Conference on Learning Representations
  (2018)

\bibitem{bengio2015scheduled}
Bengio, S., Vinyals, O., Jaitly, N., Shazeer, N.: Scheduled sampling for
  sequence prediction with recurrent neural networks. In: Advances in Neural
  Information Processing Systems. pp. 1171--1179 (2015)

\bibitem{bertalmio2000image}
Bertalmio, M., Sapiro, G., Caselles, V., Ballester, C.: Image inpainting. In:
  Proceedings of the 27th annual conference on Computer graphics and
  interactive techniques. pp. 417--424. ACM Press/Addison-Wesley Publishing Co.
  (2000)

\bibitem{brock2018large}
Brock, A., Donahue, J., Simonyan, K.: Large scale gan training for high
  fidelity natural image synthesis. arXiv preprint arXiv:1809.11096  (2018)

\bibitem{chen2017pixelsnail}
Chen, X., Mishra, N., Rohaninejad, M., Abbeel, P.: Pixelsnail: An improved
  autoregressive generative model. arXiv preprint arXiv:1712.09763  (2017)

\bibitem{choi2017stargan}
Choi, Y., Choi, M., Kim, M., Ha, J.W., Kim, S., Choo, J.: Stargan: Unified
  generative adversarial networks for multi-domain image-to-image translation.
  In: CVPR (June 2018)

\bibitem{chorowski2019unsupervised}
Chorowski, J., Weiss, R.J., Bengio, S., van~den Oord, A.: Unsupervised speech
  representation learning using wavenet autoencoders. IEEE/ACM transactions on
  audio, speech, and language processing  \textbf{27}(12),  2041--2053 (2019)

\bibitem{criminisi2003object}
Criminisi, A., Perez, P., Toyama, K.: Object removal by exemplar-based
  inpainting. In: CVPR. vol.~2, pp. II--II (2003)

\bibitem{Fauw2019HierarchicalAI}
De~Fauw, J., Dieleman, S., Simonyan, K.: Hierarchical autoregressive image
  models with auxiliary decoders. arXiv preprint arXiv:1903.04933  (2019)

\bibitem{fawzi2016image}
Fawzi, A., Samulowitz, H., Turaga, D., Frossard, P.: Image inpainting through
  neural networks hallucinations. In: 2016 IEEE 12th Image, Video, and
  Multidimensional Signal Processing Workshop (IVMSP). pp.~1--5. Ieee (2016)

\bibitem{Ganin2018spiral}
Ganin, Y., Kulkarni, T., Babuschkin, I., Eslami, S., Vinyals, O.: Synthesizing
  programs for images using reinforced adversarial learning. arXiv preprint
  arXiv:1804.01118  (2018)

\bibitem{goodfellow2014generative}
Goodfellow, I., Pouget-Abadie, J., Mirza, M., Xu, B., Warde-Farley, D., Ozair,
  S., Courville, A., Bengio, Y.: Generative adversarial nets. In: NeurIPS. pp.
  2672--2680 (2014)

\bibitem{gulrajani2017improved}
Gulrajani, I., Ahmed, F., Arjovsky, M., Dumoulin, V., Courville, A.C.: Improved
  training of wasserstein gans. In: NeurIPS. pp. 5767--5777 (2017)

\bibitem{heqing_ICIP}
Heqing, Z., Ak, K.E., Kassim, A.A.: Learning cross-modal representations for
  language-based image manipulation. In: ICIP (2020)

\bibitem{heusel2017gans}
Heusel, M., Ramsauer, H., Unterthiner, T., Nessler, B., Hochreiter, S.: Gans
  trained by a two time-scale update rule converge to a local nash equilibrium.
  In: NeurIPS. pp. 6626--6637 (2017)

\bibitem{huang2019learning}
Huang, Z., Heng, W., Zhou, S.: Learning to paint with model-based deep
  reinforcement learning. arXiv preprint arXiv:1903.04411  (2019)

\bibitem{pix2pix2017}
Isola, P., Zhu, J.Y., Zhou, T., Efros, A.A.: Image-to-image translation with
  conditional adversarial networks. In: CVPR (2017)

\bibitem{karras2017progressive}
Karras, T., Aila, T., Laine, S., Lehtinen, J.: Progressive growing of gans for
  improved quality, stability, and variation. arXiv preprint arXiv:1710.10196
  (2017)

\bibitem{karras2019style}
Karras, T., Laine, S., Aila, T.: A style-based generator architecture for
  generative adversarial networks. In: Proceedings of the IEEE Conference on
  Computer Vision and Pattern Recognition. pp. 4401--4410 (2019)

\bibitem{kingma2014adam}
Kingma, D.P., Ba, J.: Adam: A method for stochastic optimization. arXiv
  preprint arXiv:1412.6980  (2014)

\bibitem{kingma2013autoencoding}
Kingma, D.P., Welling, M.: Auto-encoding variational bayes. arXiv preprint
  arXiv:1312.6114  (2013)

\bibitem{lin2019coco}
Lin, C.H., Chang, C.C., Chen, Y.S., Juan, D.C., Wei, W., Chen, H.T.: Coco-gan:
  Generation by parts via conditional coordinating. arXiv preprint
  arXiv:1904.00284  (2019)

\bibitem{liu2015faceattributes}
Liu, Z., Luo, P., Wang, X., Tang, X.: Deep learning face attributes in the
  wild. In: ICCV (December 2015)

\bibitem{mirza2014conditional}
Mirza, M., Osindero, S.: Conditional generative adversarial nets.
  arXiv:1411.1784  (2014)

\bibitem{van2017neural}
van~den Oord, A., Vinyals, O., et~al.: Neural discrete representation learning.
  In: Advances in Neural Information Processing Systems. pp. 6306--6315 (2017)

\bibitem{oord2016pixel}
Oord, A.v.d., Kalchbrenner, N., Kavukcuoglu, K.: Pixel recurrent neural
  networks. arXiv preprint arXiv:1601.06759  (2016)

\bibitem{oord2016conditional}
Oord, A.v.d., Kalchbrenner, N., Vinyals, O., Espeholt, L., Graves, A.,
  Kavukcuoglu, K.: Conditional image generation with pixelcnn decoders. In:
  NeurIPS. pp. 4797--4805 (2016)

\bibitem{papineni2002bleu}
Papineni, K., Roukos, S., Ward, T., Zhu, W.J.: Bleu: a method for automatic
  evaluation of machine translation. In: Proceedings of the 40th annual meeting
  on association for computational linguistics. pp. 311--318. Association for
  Computational Linguistics (2002)

\bibitem{ramachandran2017fast}
Ramachandran, P., Paine, T.L., Khorrami, P., Babaeizadeh, M., Chang, S., Zhang,
  Y., Hasegawa-Johnson, M.A., Campbell, R.H., Huang, T.S.: Fast generation for
  convolutional autoregressive models. arXiv preprint arXiv:1704.06001  (2017)

\bibitem{Ranzato2015SequenceLT}
Ranzato, M., Chopra, S., Auli, M., Zaremba, W.: Sequence level training with
  recurrent neural networks. arXiv preprint arXiv:1511.06732  (2015)

\bibitem{razavi2019generating}
Razavi, A., Oord, A.v.d., Vinyals, O.: Generating diverse high-fidelity images
  with vq-vae-2. arXiv preprint arXiv:1906.00446  (2019)

\bibitem{saito2017statistical}
Saito, Y., Takamichi, S., Saruwatari, H.: Statistical parametric speech
  synthesis incorporating generative adversarial networks. IEEE/ACM
  \textbf{26}(1),  84--96 (2017)

\bibitem{salimans2016improved}
Salimans, T., Goodfellow, I., Zaremba, W., Cheung, V., Radford, A., Chen, X.:
  Improved techniques for training gans. In: NeurIPS. pp. 2234--2242 (2016)

\bibitem{sisman2019singan}
{Sisman}, B., {Vijayan}, K., {Dong}, M., {Li}, H.: Singan: Singing voice
  conversion with generative adversarial networks. In: APSIPA ASC. pp. 112--118
  (2019)

\bibitem{sisman2020generative}
Sisman, B., Li, H.: Generative adversarial networks for singing voice
  conversion with and without parallel data. In: Speaker Odyssey. pp. 238--244
  (2020)

\bibitem{sisman2019study}
Sisman, B., Zhang, M., Dong, M., Li, H.: On the study of generative adversarial
  networks for cross-lingual voice conversion. In: 2019 IEEE Automatic Speech
  Recognition and Understanding Workshop (ASRU). pp. 144--151. IEEE (2019)

\bibitem{sisman2018adaptive}
Sisman, B., Zhang, M., Sakti, S., Li, H., Nakamura, S.: Adaptive wavenet
  vocoder for residual compensation in gan-based voice conversion. In: SLT. pp.
  282--289 (2018)

\bibitem{SutskeverVL14}
Sutskever, I., Vinyals, O., Le, Q.V.: Sequence to sequence learning with neural
  networks. In: NeurIPS. pp. 3104--3112 (2014)

\bibitem{sutton2000policy}
Sutton, R.S., McAllester, D.A., Singh, S.P., Mansour, Y.: Policy gradient
  methods for reinforcement learning with function approximation. In: Advances
  in neural information processing systems. pp. 1057--1063 (2000)

\bibitem{theis2015note}
Theis, L., Oord, A.v.d., Bethge, M.: A note on the evaluation of generative
  models. arXiv preprint arXiv:1511.01844  (2015)

\bibitem{tjandra2019vqvae}
Tjandra, A., Sisman, B., Zhang, M., Sakti, S., Li, H., Nakamura, S.: Vqvae
  unsupervised unit discovery and multi-scale code2spec inverter for zerospeech
  challenge 2019. arXiv preprint arXiv:1905.11449  (2019)

\bibitem{toyama2018toward}
Toyama, J., Iwasawa, Y., Nakayama, K., Matsuo, Y.: Toward learning better
  metrics for sequence generation training with policy gradient  (2018)

\bibitem{williams1992simple}
Williams, R.J.: Simple statistical gradient-following algorithms for
  connectionist reinforcement learning. Machine learning  \textbf{8}(3-4),
  229--256 (1992)

\bibitem{Tao18attngan}
Xu, T., Zhang, P., Huang, Q., Zhang, H., Gan, Z., Huang, X., He, X.: Attngan:
  Fine-grained text to image generation with attentional generative adversarial
  networks. In: {CVPR} (2018)

\bibitem{yu15lsun}
Yu, F., Zhang, Y., Song, S., Seff, A., Xiao, J.: Lsun: Construction of a
  large-scale image dataset using deep learning with humans in the loop. arXiv
  preprint arXiv:1506.03365  (2015)

\bibitem{yu2019free}
Yu, J., Lin, Z., Yang, J., Shen, X., Lu, X., Huang, T.S.: Free-form image
  inpainting with gated convolution. In: ICCV. pp. 4471--4480 (2019)

\bibitem{yu2016seqgan}
Yu, L., Zhang, W., Wang, J., Yu, Y.: Seqgan: Sequence generative adversarial
  nets with policy gradient. In: AAAI (2017)

\bibitem{zhang2018self}
Zhang, H., Goodfellow, I., Metaxas, D., Odena, A.: Self-attention generative
  adversarial networks. arXiv preprint arXiv:1805.08318  (2018)

\bibitem{zhou2020transforming}
Zhou, K., Sisman, B., Li, H.: Transforming spectrum and prosody for emotional
  voice conversion with non-parallel training data. arXiv preprint
  arXiv:2002.00198  (2020)

\end{thebibliography}


\begin{thebibliography}{1}
\providecommand{\url}[1]{\texttt{#1}}
\providecommand{\urlprefix}{URL }
\providecommand{\doi}[1]{https://doi.org/#1}

\bibitem{karras2017progressive}
Karras, T., Aila, T., Laine, S., Lehtinen, J.: Progressive growing of gans for
  improved quality, stability, and variation. arXiv preprint arXiv:1710.10196
  (2017)

\bibitem{liu2015faceattributes}
Liu, Z., Luo, P., Wang, X., Tang, X.: Deep learning face attributes in the
  wild. In: ICCV (December 2015)

\bibitem{van2017neural}
van~den Oord, A., Vinyals, O., et~al.: Neural discrete representation learning.
  In: Advances in Neural Information Processing Systems. pp. 6306--6315 (2017)

\bibitem{oord2016conditional}
Oord, A.v.d., Kalchbrenner, N., Vinyals, O., Espeholt, L., Graves, A.,
  Kavukcuoglu, K.: Conditional image generation with pixelcnn decoders. In:
  NeurIPS. pp. 4797--4805 (2016)

\bibitem{yu15lsun}
Yu, F., Zhang, Y., Song, S., Seff, A., Xiao, J.: Lsun: Construction of a
  large-scale image dataset using deep learning with humans in the loop. arXiv
  preprint arXiv:1506.03365  (2015)

\end{thebibliography}
\end{document}


\title{Incorporating Reinforced Adversarial Learning in Autoregressive Image Generation \\  -Supplementary Material-}

\maketitle

\begin{figure*}
\centering
\includegraphics[scale=0.46]{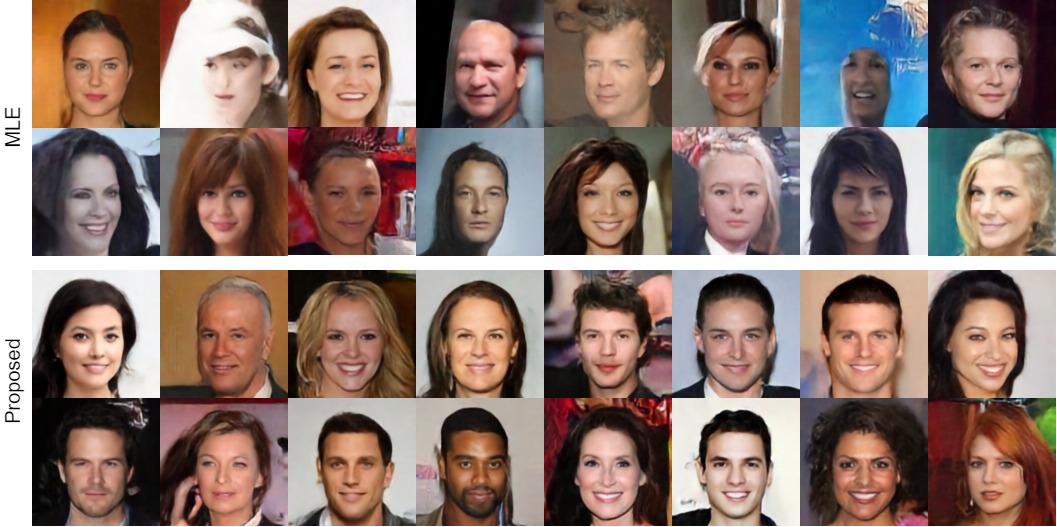}
\caption{Additional image samples generated by MLE-trained model and our proposed reinforced adversarial learning on the CelebA dataset. The generated images have resolution of $128  \times 128$, and they are decoded from $16 \times 16$ and $32 \times 32$ latent codes.}
\label{fig:celeba_128}
\end{figure*}

\begin{figure*}
\centering
\includegraphics[scale=0.46]{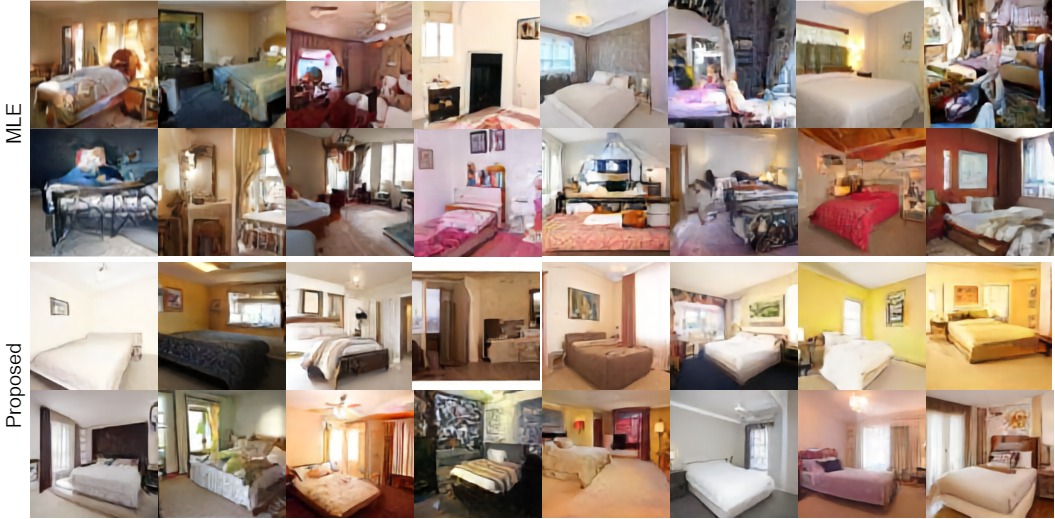}
\caption{Additional image samples generated by MLE-trained model and our proposed reinforced adversarial learning on the LSUN-bedroom dataset. The generated images have resolution of $128  \times 128$, and they are decoded from $16 \times 16$ and $32 \times 32$ latent codes.}
\label{fig:lsun_128}
\end{figure*}

\section{Additional Image Samples}
We provide additional image samples in Figure \ref{fig:celeba_128} and Figure \ref{fig:lsun_128} with the same models used in Figure 5 and Figure 6 of the main paper. The images have resolution of $128 \times 128$, decoded from latent code sizes of $16 \times 16$ and $32 \times 32$.

We further provide quantitative results in Table \ref{table:celeba} for CelebA \cite{liu2015faceattributes} and in Table \ref{table:lsun} for LSUN-bedroom \cite{yu15lsun} datasets with different settings. Lastly, we also share an experiment on CelebA-HQ \cite{karras2017progressive} dataset for image resolution of $512 \times 512$ in Table \ref{table:celeba_HQ} using latent code sizes of $32 \times 32$ and $64 \times 64$. For all experiments, we finetune the pixelcnn model with the MLE loss with the same number of iterations as our RL trained model, then compare which one is better.

\section{Architecture Details}
We provide the details of PixelCNN \cite{oord2016conditional} and VQVAE \cite{van2017neural} architectures we used in our experiments in Tables \ref{table:celeba_64_oracle}, \ref{table:arch_lsun_64_8}, \ref{table:lsun_64_16}, \ref{table:celeb_lsun_128_single}. \ref{table:celeb_lsun_128_double}, \ref{table:celeba_256_double} and \ref{table:celeb_HQ_double}. 
\section{Sampling Speed}
We report FID for a partial generation and full generation under the same training time in the main paper, Table 3(a). More detailed training durations are reported in Table \ref{table:sampling_speed}, computed with a single GPU. For  $128 \times 128$ res., partial generation can sample a batch of 16 images in 61 seconds on average compared to 102 seconds with full generation. For $ 256\times 256$  res., partial generation can sample a batch of 16 images in 123 seconds on average compared to 170 seconds with full generation. As can be seen, the partial generation significantly improves the training time. Due to smaller code sizes, the model is also less exhaustive to train as lesser GPU memory is allocated.

\section{Scaling to Higher Resolutions}
The main reason we do not have such successful results is due to computational limitations as discussed in the paper. Another reason is the VQ-VAE reconstruction and even calculating FIDs for real images vs. reconstructed real images, FIDs are very high as can be seen in Table \ref{table:FID_recons}. As we limit the encoded code size to effectively use PixelCNNs, FIDs reduce due to the increased compression. It is noteworthy that the ImageNet classifier is also very sensitive to small distortions caused by vector quantization as also discussed in VQ-VAE-2.

The superior performance of VQ-VAE-2 is also enabled by rejection sampling. Rejection sampling is an external method to filter out bad quality samples by using a classifier network that is trained on ImageNet and can be applied to ANY generative network. However, the rejection sampling method is a temporary solution, which makes it hard to use for real-world cases. By introducing Reinforced Adversarial Learning, our aim is to penalize bad quality samples and permanently improve PixenCNNs sample generation quality.

Another issue with rejection sampling is sampling costs. The sampling time for autoregressive models is already a problem and including rejection sampling would make this process even slower. For $256 \times 256$ resolution, as reported in VQ-VAE-2 supplementary materials, the sampling speed per image is 3 minutes even with caching and use of large batches. Therefore, VQ-VAE-2 requires many GPU hours, which is very costly. In the paper, we mention the disadvantages of rejection sampling such as increased sampling time, reduced diversity and unavailability of class scores.

\begin{table*}
\caption{Additional experiment for CelebA dataset. FID calculations are made compared to reconstructed real images.} \label{table:celeba}
\centering
\begin{tabular}{|c|c|c|}
\hline
Image Size | Latent Code Size   & $128 \times 128$ | $16\times 16$  \\ \hline
MLE &  18.01 \\ \hline
Single Reward & 16.91 \\ \hline
\end{tabular}
\label{table:FID_real}
\end{table*}

\begin{table*}
\caption{Additional experiments for LSUN-bedroom dataset. FID calculations are made compared to reconstructed real images.} \label{table:lsun}
\centering
\begin{tabular}{|c|c|c|c|c|}
\hline
Image Size | Latent Code Size & $64 \times 64$ | $8\times 8$ & $64 \times 64$ | $16\times 16$ & $128 \times 128$ | $16\times 16$  \\ \hline
MLE & 19.95 & 15.90 & 23.53 \\ \hline
Single Reward & 19.07 & 11.47 & 18.96 \\ \hline
\end{tabular}
\label{table:FID_real}
\end{table*}

\begin{table*}
\caption{Additional experiment for CelebA-HQ dataset. FID calculations are made compared to reconstructed real images.} \label{table:celeba_HQ}
\centering
\begin{tabular}{|c|c|c|}
\hline
Image Size | Latent Code Size   & $512 \times 512$ | top: $32\times 32$, bottom: $64\times 64$  \\ \hline
MLE &  34.93 \\ \hline
Single Reward & 32.14 \\ \hline
Intermediate Reward &  31.84\\ \hline
\end{tabular}
\end{table*}

\begin{table}[!htb]
\caption{Architecture details for CelebA, $64 \times 64$ image resolution experiments.}
\label{table:celeba_64_oracle}
\begin{subtable}{.35\linewidth}
\centering
\caption{PixelCNN Prior - Oracle Network}
\begin{tabular}{cc}
\hline
Input size &  $8 \times 8$ \\ \hline
Batch size & 1024 \\ \hline
Hidden units & 256 \\ \hline 
Residual units & 256 \\ \hline 
Layers & 5 \\ \hline 
Attention layers & 0 \\ \hline 
Conv Filter size & 5 \\ \hline 
Dropout & 0.1 \\ \hline 
Output stack layers & 5 \\ \hline 
Training steps & 90039 \\ \hline 
\end{tabular}
\end{subtable}%
\begin{subtable}{.35\linewidth}
\centering
\caption{PixelCNN Prior - Generator Network}
\begin{tabular}{cc}
\hline
Input size & $8 \times 8$  \\ \hline
Batch size & 1024 \\ \hline
Hidden units & 128 \\ \hline 
Residual units & 128 \\ \hline 
Layers &  5 \\ \hline 
Attention layers & 0 \\ \hline 
Conv Filter size & 5 \\ \hline 
Dropout &  0.1 \\ \hline 
Output stack layers & 5 \\ \hline
Training steps & 19531 \\ \hline 
\end{tabular}
\end{subtable}%
\begin{subtable}{.32\linewidth}
\centering
\caption{VQ-VAE: Encoder, Decoder}
\begin{tabular}{cc}
\hline
Input size &  $64 \times 64$ \\ \hline
Latent Layers & $8 \times 8$  \\ \hline
$\beta$  & 0.25 \\ \hline
Batch size & 128 \\ \hline
Hidden units & 128 \\ \hline 
Residual units &  64 \\ \hline 
Layers &  3 \\ \hline 
Codebook size & 128 \\ \hline 
Codebook dimension & 64 \\ \hline 
Encoder filter size & 3 \\ \hline 
Decoder filter size & 4 \\ \hline 
Training steps & 156250 \\ \hline 
\end{tabular}
\end{subtable} 
\end{table}

\begin{table}[!htb]
\caption{Architecture details for LSUN-bedroom, $64 \times 64$ image resolution experiments with latent size of  $8 \times 8$.}
\label{table:arch_lsun_64_8}
\begin{subtable}{.5\linewidth}
\centering
\caption{PixelCNN Prior}
\begin{tabular}{cc}
\hline
Input size &  $8 \times 8$ \\ \hline
Batch size & 1536 \\ \hline
Hidden units & 512 \\ \hline 
Residual units & 512 \\ \hline 
Layers & 20 \\ \hline 
Attention layers & 0 \\ \hline 
Conv Filter size & 5 \\ \hline 
Dropout & 0.1 \\ \hline 
Output stack layers & 5 \\ \hline 
Training steps & 97656 \\ \hline 
\end{tabular}
\end{subtable}%
\begin{subtable}{.5\linewidth}
\centering
\caption{VQ-VAE: Encoder, Decoder}
\begin{tabular}{cc}
\hline
Input size &  $64 \times 64$ \\ \hline
Latent Layers & $8 \times 8$  \\ \hline
$\beta$  & 0.25 \\ \hline
Batch size & 128 \\ \hline
Hidden units & 128 \\ \hline 
Residual units &  64 \\ \hline 
Layers &  3 \\ \hline 
Codebook size & 512 \\ \hline 
Codebook dimension & 64 \\ \hline 
Encoder filter size & 3 \\ \hline 
Decoder filter size & 4 \\ \hline 
Training steps & 234375 \\ \hline 
\end{tabular}
\end{subtable} 
\end{table}

\begin{table}[!htb]
\caption{Architecture details for LSUN-bedroom, $64 \times 64$ image resolution experiments with latent size of  $16 \times 16$.}
\label{table:lsun_64_16}
\begin{subtable}{.5\linewidth}
\centering
\caption{PixelCNN Prior}
\begin{tabular}{cc}
\hline
Input size &  $16 \times 16$ \\ \hline
Batch size & 480 \\ \hline
Hidden units & 512 \\ \hline 
Residual units & 512 \\ \hline 
Layers & 20 \\ \hline 
Attention layers & 8 \\ \hline 
Conv Filter size & 5 \\ \hline 
Dropout & 0.1 \\ \hline 
Output stack layers & 20 \\ \hline 
Training steps & 31250 \\ \hline 
\end{tabular}
\end{subtable}%
\begin{subtable}{.5\linewidth}
\centering
\caption{VQ-VAE: Encoder, Decoder}
\begin{tabular}{cc}
\hline
Input size &  $64 \times 64$ \\ \hline
Latent Layers & $16 \times 16$  \\ \hline
$\beta$  & 0.25 \\ \hline
Batch size & 128 \\ \hline
Hidden units & 128 \\ \hline 
Residual units &  64 \\ \hline 
Layers &  2 \\ \hline 
Codebook size & 512 \\ \hline 
Codebook dimension & 64 \\ \hline 
Encoder filter size & 3 \\ \hline 
Decoder filter size & 4 \\ \hline 
Training steps & 234375 \\ \hline 
\end{tabular}
\end{subtable} 
\end{table}

\begin{table}[!htb]
\caption{Architecture details for CelebA and LSUN-bedroom with $128 \times 128$ resolution with a single prior PixelCNNs.}
\label{table:celeb_lsun_128_single}
\begin{subtable}{.5\linewidth}
\centering
\caption{PixelCNN Prior}
\begin{tabular}{cc}
\hline
Input size &  $16 \times 16$ \\ \hline
Batch size & 480 \\ \hline
Hidden units & 512 \\ \hline 
Residual units & 512 \\ \hline 
Layers & 20 \\ \hline 
Attention layers & 8 \\ \hline 
Conv Filter size & 5 \\ \hline 
Dropout & 0.1 \\ \hline 
Output stack layers & 20 \\ \hline 
Training steps & 41666 \\ \hline 
\end{tabular}
\end{subtable}%
\begin{subtable}{.5\linewidth}
\centering
\caption{VQ-VAE: Encoder, Decoder}
\begin{tabular}{cc}
\hline
Input size &  $128 \times 128$ \\ \hline
Latent Layers & $16 \times 16$  \\ \hline
$\beta$  & 0.25 \\ \hline
Batch size & 128 \\ \hline
Hidden units & 128 \\ \hline 
Residual units &  64 \\ \hline 
Layers &  3 \\ \hline 
Codebook size & 512 \\ \hline 
Codebook dimension & 64 \\ \hline 
Encoder filter size & 3 \\ \hline 
Decoder filter size & 4 \\ \hline 
Training steps & 156250 \\ \hline 
\end{tabular}
\end{subtable} 
\end{table}

\begin{table}[!htb]
\caption{Architecture details for CelebA,  LSUN-bedroom, $128 \times 128$ experiments with two PixelCNN priors.}
\label{table:celeb_lsun_128_double}

\begin{subtable}{.5\linewidth}
\centering
\caption{PixelCNN Top-Prior Network}
\begin{tabular}{cc}
\hline
Input size &  $16 \times 16$ \\ \hline
Batch size & 480 \\ \hline
Hidden units & 512 \\ \hline 
Residual units & 512 \\ \hline 
Layers & 20 \\ \hline 
Attention layers & 8 \\ \hline 
Conv Filter size & 5 \\ \hline 
Dropout & 0.1 \\ \hline 
Output stack layers & 20 \\ \hline 
Training steps & 166664 \\ \hline 
\end{tabular}
\end{subtable}%
\begin{subtable}{.5\linewidth}
\centering
\caption{PixelCNN Bottom-Prior Network}
\begin{tabular}{ll}
\hline
Input size &  $32 \times 32$ \\ \hline
Batch size & 128 \\ \hline
Hidden units & 512 \\ \hline 
Residual units & 512 \\ \hline 
Layers & 20 \\ \hline 
Attention layers & 0 \\ \hline 
Conv Filter size & 5 \\ \hline 
Dropout & 0.1 \\ \hline 
Conditioning Output stack layers & 20 \\ \hline 
Training steps & 156250 \\ \hline 
\end{tabular}
\end{subtable}%

\begin{subtable}{1\linewidth}
\centering
\caption{VQ-VAE: Encoder, Decoder}
\begin{tabular}{cc}
\hline
Input size &  $128 \times 128$ \\ \hline
Latent Layers & $16 \times 16$, $32 \times 32$  \\ \hline
$\beta$  & 0.25 \\ \hline
Batch size & 128 \\ \hline
Hidden units & 128 \\ \hline 
Residual units &  64 \\ \hline 
Layers &  2 \\ \hline 
Codebook size & 512 \\ \hline 
Codebook dimension & 64 \\ \hline 
Encoder filter size & 3 \\ \hline 
Decoder filter size & 4 \\ \hline 
Training steps & 156250 \\ \hline 
\end{tabular}
\end{subtable} 
\end{table}

\begin{table}[!htb]
\caption{Architecture details for CelebA, $256 \times 256$ experiments with two PixelCNN priors.}
\label{table:celeba_256_double}
\begin{subtable}{.5\linewidth}
\centering
\caption{PixelCNN Top-Prior Network}
\begin{tabular}{cc}
\hline
Input size &  $32 \times 32$ \\ \hline
Batch size & 128 \\ \hline
Hidden units & 512 \\ \hline 
Residual units & 512 \\ \hline 
Layers & 20 \\ \hline 
Attention layers & 0 \\ \hline 
Conv Filter size & 5 \\ \hline 
Dropout & 0.1 \\ \hline 
Output stack layers & 20 \\ \hline 
Training steps & 187500 \\ \hline 
\end{tabular}
\end{subtable}%
\begin{subtable}{.5\linewidth}
\centering
\caption{PixelCNN Bottom-Prior Network}
\begin{tabular}{ll}
\hline
Input size &  $64 \times 64$ \\ \hline
Batch size & 128 \\ \hline
Hidden units & 128 \\ \hline 
Residual units & 128 \\ \hline 
Layers & 16 \\ \hline 
Attention layers & 0 \\ \hline 
Conv Filter size & 5 \\ \hline 
Dropout & 0.1 \\ \hline 
Conditioning Output stack layers & 20 \\ \hline 
Training steps & 125000 \\ \hline 
\end{tabular}
\end{subtable}%

\begin{subtable}{1\linewidth}
\centering
\caption{VQ-VAE: Encoder, Decoder}
\begin{tabular}{cc}
\hline
Input size &  $256 \times 256$ \\ \hline
Latent Layers & $32 \times 32$, $64 \times 64$  \\ \hline
$\beta$  & 0.25 \\ \hline
Batch size & 128 \\ \hline
Hidden units & 128 \\ \hline 
Residual units &  64 \\ \hline 
Layers &  2 \\ \hline 
Codebook size & 512 \\ \hline 
Codebook dimension & 64 \\ \hline 
Encoder filter size & 3 \\ \hline 
Decoder filter size & 4 \\ \hline 
Training steps & 156250 \\ \hline 
\end{tabular}
\end{subtable} 
\end{table}

\begin{table}[!htb]
\caption{Architecture details for CelebA-HQ, $512 \times 512$ experiments with two PixelCNN priors.}
\label{table:celeb_HQ_double}
\begin{subtable}{.5\linewidth}
\centering
\caption{PixelCNN Top-Prior Network}
\begin{tabular}{cc}
\hline
Input size &  $32 \times 32$ \\ \hline
Batch size & 128 \\ \hline
Hidden units & 512 \\ \hline 
Residual units & 512 \\ \hline 
Layers & 20 \\ \hline 
Attention layers & 0 \\ \hline 
Conv Filter size & 5 \\ \hline 
Dropout & 0.1 \\ \hline 
Output stack layers & 20 \\ \hline 
Training steps & 20625 \\ \hline 
\end{tabular}
\end{subtable}%
\begin{subtable}{.5\linewidth}
\centering
\caption{PixelCNN Bottom-Prior Network}
\begin{tabular}{ll}
\hline
Input size &  $64 \times 64$ \\ \hline
Batch size & 128 \\ \hline
Hidden units & 128 \\ \hline 
Residual units & 128 \\ \hline 
Layers & 16 \\ \hline 
Attention layers & 0 \\ \hline 
Conv Filter size & 5 \\ \hline 
Dropout & 0.1 \\ \hline 
Conditioning Output stack layers & 20 \\ \hline 
Training steps & 42187 \\ \hline 
\end{tabular}
\end{subtable}%

\begin{subtable}{1\linewidth}
\centering
\caption{VQ-VAE: Encoder, Decoder}
\begin{tabular}{cc}
\hline
Input size &  $512 \times 512$ \\ \hline
Latent Layers & $32 \times 32$, $64 \times 64$  \\ \hline
$\beta$  & 0.25 \\ \hline
Batch size & 128 \\ \hline
Hidden units & 128 \\ \hline 
Residual units &  64 \\ \hline 
Layers &  3 \\ \hline 
Codebook size & 512 \\ \hline 
Codebook dimension & 64 \\ \hline 
Encoder filter size & 3 \\ \hline 
Decoder filter size & 4 \\ \hline 
Training steps & 23437 \\ \hline 
\end{tabular}
\end{subtable} 
\end{table}

\begin{table*}
\caption{Sampling speed comparisons for full generation vs. partial generation with a batch size of 16.} \label{table:sampling_speed}
\centering
\begin{tabular}{|c|c|c|c|}
\hline
Image Size | Latent Code Size & $128 \times 128$ | $16\times 16$ + $32\times 32$ & $256 \times 256$ | $32\times 32$ + $64\times 64$   \\ \hline
Full Generation & 102 s & 170 s  \\ \hline
Partial Generation  & 61 s & 123 s  \\ \hline
\end{tabular}
\end{table*}

\begin{table*}
\caption{FIDs for real images vs. reconstructed real images.} \label{table:FID_recons}
\begin{tabular}{|c|c|c|c|}
\hline
\begin{tabular}[c]{@{}c@{}}Dataset\\ Image Size\\ Latent Code Size\end{tabular} & \begin{tabular}[c]{@{}c@{}}Celeba\\ 128 x 128\\ 16 x 16 + 32 x 32\end{tabular} & \begin{tabular}[c]{@{}c@{}}Celeba \\ 256 x 256 \\ 32 x 32 + 64 x 64\end{tabular} & \begin{tabular}[c]{@{}c@{}}LSUN-bedroom\\ 128 x 128\\ 16 x 16 + 32 x 32\end{tabular} \\ \hline
FID: Real image vs. Reconstructed Real Image                                                                             & 24.52                                                                          & 7.47                                                                             & 28.54                                                                                \\ \hline
\end{tabular}
\end{table*}

\clearpage
%
%
\bibliographystyle{splncs04}
\bibliography{MyCollection}